\documentclass[9pt,journal,compsoc]{IEEEtran}
\ifCLASSOPTIONcompsoc
  \usepackage[nocompress]{cite}
\else
  \usepackage{cite}
\fi

\usepackage{flushend}

\ifCLASSINFOpdf
  \usepackage[pdftex]{graphicx}
  \graphicspath{{./}}
  \DeclareGraphicsExtensions{.jpg,.png,.pdf,.svg,.eps}
\else
  \usepackage[dvips]{graphicx}
\fi

\usepackage{array,multirow}

\usepackage{url}

\usepackage{listings}
\usepackage{color}

\definecolor{codegreen}{rgb}{0,0.6,0}
\definecolor{codegray}{rgb}{0.5,0.5,0.5}
\definecolor{codepurple}{rgb}{0.58,0,0.82}
\definecolor{backcolour}{rgb}{1,1,1}

\lstdefinestyle{mystyle}{
	backgroundcolor=\color{backcolour},   
	commentstyle=\color{codegreen},
	keywordstyle=\bfseries\color{black},
	numberstyle=\tiny\color{codegray},
	stringstyle=\color{codepurple},
	basicstyle=\footnotesize,
	breakatwhitespace=false,         
	breaklines=true,                 
	captionpos=b,                    
	keepspaces=true,                 
	numbers=left,                    
	numbersep=5pt,                  
	showspaces=false,                
	showstringspaces=false,
	showtabs=false,                  
	tabsize=2
}

\usepackage[acronym]{glossaries}
\makeglossaries

\newacronym{cpu}{CPU}{Central Processing Unit}
\newacronym{cnn}{CNN}{Convolutional Neural Network}
\newacronym{mac}{MAC}{Multiply Accumulate}
\newacronym{asic}{ASIC}{Application Specific Integrated Circuit}
\newacronym{ieee}{IEEE}{Institute of Electrical and Electronic Engineers}
\newacronym{ic}{IC}{Integrated Circuit}
\newacronym{fpga}{FPGA}{Field Programmable Gate Array}
\newacronym{dsp}{DSP}{Digital Signal Processor}
\newacronym{bram}{BRAM}{Block RAM}
\newacronym{relu}{ReLU}{Rectified Linear Unit}
\newacronym{gpu}{GPU}{Graphics Processor Unit}
\newacronym{pas}{PAS}{Parallel Accumulate and Store}
\newacronym{pasm}{PASM}{Parallel Accumulate Shared MAC}
\newacronym{rtl}{RTL}{Register Transfer Level}
\newacronym{ip}{IP}{Intellectual Property}
\newacronym{hdl}{HDL}{Hardware Description Language} 
\newacronym{sdc}{SDC}{Synopsys Design Constraints}
\newacronym{xdc}{XDC}{Xilinx Design Constraint}
\newacronym{iob}{IOB}{Input Output Buffer}
\newacronym{ram}{RAM}{Random Access Memory}
\newacronym{xpe}{XPE}{Xilinx Power Estimator}
\newacronym{clb}{CLB}{Configurable Logic Block}
\newacronym{lut}{LUT}{Look Up Table}
\newacronym{mux}{MUX}{Multiplexer}
\newacronym{sop}{SOP}{Sum Of Products}

\begin{document}
\title{Low Complexity \acrlong{mac}\\Unit for Weight-Sharing \\\acrlong{cnn}s}

\author{James~Garland and David~Gregg, Trinity College Dublin
\IEEEcompsocitemizethanks{\IEEEcompsocthanksitem Manuscript
submitted: 30-Aug-2016.  Manuscript accepted: 14-Dec-2016.  Final manuscript
received: 19-Jan-2016.
\IEEEcompsocthanksitem This research is supported by Science Foundation Ireland, Project 12/IA/1381.}}

\IEEEtitleabstractindextext{
\begin{abstract}
\acrlong{cnn}s (\glspl{cnn}) are one of the most successful deep machine learning technologies for processing image, voice and video data. \glspl{cnn} require large amounts of processing capacity and memory, which can exceed the resources of low power mobile and embedded systems. Several designs for hardware accelerators have been proposed for \glspl{cnn} which typically contain large numbers of \acrfull{mac} units. One approach to reducing data sizes and memory traffic in \gls{cnn} accelerators is ``weight sharing'', where the full range of values in a trained \gls{cnn} are put in bins and the bin index is stored instead of the original weight value. In this paper we propose a novel \gls{mac} circuit that exploits binning in weight-sharing \glspl{cnn}. Rather than computing the \gls{mac} directly we instead count the frequency of each weight and place it in a bin. We then compute the accumulated value in a subsequent multiply phase. This allows hardware multipliers in the \gls{mac} circuit to be replaced with adders and selection logic. Experiments show that for the same clock speed our approach results in fewer gates, smaller logic, and reduced power.
\end{abstract}

}

\maketitle

\IEEEraisesectionheading
{\section{Introduction}
\label{sec:introduction}}
\IEEEPARstart{C}{onvolutional} neural networks require large amounts of computation and weight data that stretch the limited battery, computation and memory in mobile systems. Researchers have proposed hardware accelerators for \glspl{cnn} that include many parallel hardware multiply-accumulate units.

\gls{cnn} training involves incrementally modifying the numeric ``weight'' values associated with connections in the neural network. Once training is complete, there are no further updates of the weight values, and the trained network can be deployed to large numbers of embedded devices. Han \MakeLowercase{\textit{et al.}} \cite{DeepCompression:Han,eie:Han} found that similar weight values can occur multiple times in a trained \glspl{cnn}. By binning the weights and retraining the network with the binned values, they found that just 16 weights were sufficient in many cases. They encode the weights by replacing the original numeric values with a four-bit index that specifies which of the 16 shared weights should be used. This greatly reduces the size of the weight matrices.

We propose a novel \acrfull{mac} circuit design aimed specifically at hardware accelerators for \glspl{cnn} that use weight sharing. Rather than performing the \gls{mac} operation directly, we instead count the frequency of each weight and accumulate the corresponding image value in a bin. This allows the hardware multiplier in the \gls{mac} to be replaced with counting and selection logic. After accumulation, the accumulated image values in the bins are then multiplied with the corresponding weight value of that bin. Where the number of bins is small, the counting and selection logic can be significantly smaller than the corresponding multiplier.

\section{Background}
\label{subsec:background}
The most computationally-intensive part of a \gls{cnn} is the multi-channel convolution. Fig. \ref{fig:covCode} shows pseudo-code for the main loop of the 2D multi-channel convolution operator. The input of the operator consists of an image with many (often 256 or 512 \cite{GoingDeeperWithConv:szegedy}) channels, which is combined with a large number of kernels to create an output image, again with many channels. The innermost loops can be implemented with a hardware \gls{mac} and accelerators for \glspl{cnn} commonly include a large number of parallel \glspl{mac} \cite{OptimizingFpgaAccelForCNN:Zhang}.

\begin{figure}
\begin{minipage}{1\linewidth}
\begin{lstlisting}[language=VHDL, basicstyle=\footnotesize, breaklines=true]
input[width][height][input_channels];
kernels[output_channels][K][K][input_channels];
output[width][height][output_channels];
for w in 1 to width
  for h in 1 to height
    for o in 1 to output_channels {
      sum = 0;
      for x in 1 to K
        for y in 1 to K
          for i in 1 to input_channels
            sum += input[w+x][h+y][i] *
                           kernels[o][x][y][i];
      output[w][h][o] = sum;
    }
\end{lstlisting}
\end{minipage}
	\setlength{\abovecaptionskip}{-0.25cm}
    \caption{Simplified code for 2D multi-channel convolution with a single multi-channel input and multiple multi-channel convolution kernels. Note that special treatment of edge boundaries is not shown in this code.}
    \label{fig:covCode}
\end{figure}

\begin{figure}[t]
	\centering
	\includegraphics[width=1\linewidth,keepaspectratio]{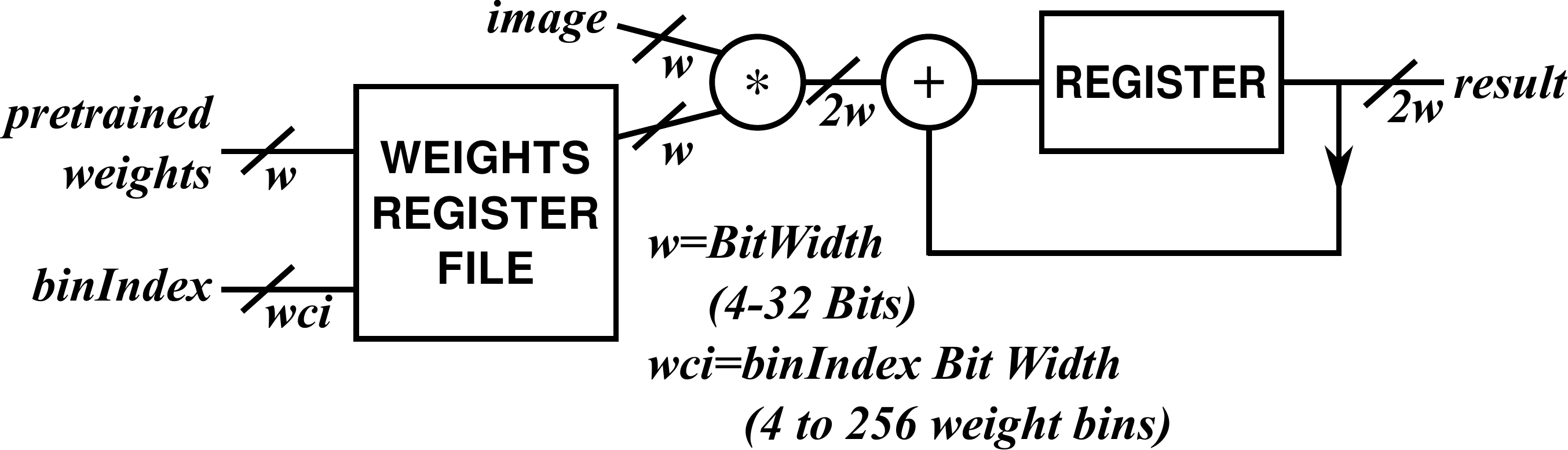}
	\setlength{\abovecaptionskip}{-0.25cm}
	\caption{Simple \gls{mac} showing how pre-trained \textbf{\textit{weight}} values are stored in a weights register file. Values are indexed and retrieved by \textbf{\textit{binIndex}} and multiplied by the corresponding \textbf{\textit{image}} value.}
	\label{fig:weightSharedRegFileAndMaccBlockDiagram}
\end{figure}

A simple \gls{mac} unit is a sequential circuit that accepts a pair of numeric values, computes their product and accumulates the result. The accumulator register is typically located within the \gls{mac} to reduce routing complexity and delays. When one of the inputs to the \gls{mac} is encoded using weight sharing, an extra level of indirection is required. Fig. \ref{fig:weightSharedRegFileAndMaccBlockDiagram} shows a simplified version of weight sharing decode with a single \gls{mac}. The weights in the \gls{cnn} are represented by, for example, a 4-bit number which is an index to a table of 16 shared weight values.

The most common computation in a \gls{cnn} is multiply-accumulate. Hardware accelerators for \glspl{cnn} may contain dozens, hundreds or even thousands of \gls{mac} units. However, each \gls{mac} unit is a relatively expensive piece of hardware consuming large floor area (i.e. large numbers of gates) and power in an \gls{asic}. Hardware accelerators for \glspl{cnn} typically use 8-, 16-, 24- or 32-bit fixed point arithmetic \cite{Eyeriss:Chen}. A combinatorial $w$-bit multiplier requires $O(w^2)$ logic gates to implement which makes up a large part of the \gls{mac} unit. Note that sub-quadratic multipliers are possible, but are inefficient for practical values of $w$ \cite{FastIntegerMultiplication:Furer}.

The \textit{gates} column of Table \ref{table:complexity} shows the circuit complexity in gates of each sub-component, assuming fixed-point arithmetic.  The bit-width of the data is $w$ and the number of bins is $b$ in the weight-shared designs. For example, a simple \gls{mac} unit contains an adder ($O(w)$ gates) a multiplier ($O(w)^2$) gates) and a register ($O(w)$ gates). A weight shared \gls{mac} also needs a small register file with $b$ entries to allow fast mapping of encoded weight indices to shared weights. The \gls{pas} is our novel weight-shared \gls{mac} unit which is described in Section \ref{sec:approach}.

\begin{table}
	\centering
	\caption{Complexity of \gls{mac}, Weight-shared \gls{mac} and \gls{pas}}
	\begin{tabular}{l|c|c|c|c}
		\hline
		Sub Component   	& Gates            & Simple & Weight Shared & PAS \\
							&                  & MAC    & MAC			&     \\
		\hline
		Adder				& $O(w)$           & 1      & 1      		& 1   \\
		Multiplier          & $O(w^2)$         & 1      & 1      		&     \\
		Register            & $O(w)$           & 1      & $b$      		& $b$ \\
		Register			&				   &		&				& 	  \\
		File Port			& $O(wb)$          &        & 1      		& 2   \\
		\hline
	\end{tabular}
	\label{table:complexity}
\end{table}
 
\section{Approach}
\label{sec:approach}
We propose to reduce the area and power consumption of the \glspl{mac} by re-architecting the \gls{mac} to do the accumulation first followed by a shared post-pass multiplication. Rather than computing the \gls{sop} in the \gls{mac} directly, we instead count how many times each of the $b$ weight indexes appears and store the corresponding image value in a register bin. For example, if the shared weight with index $2$ had the value $19$ and were multiplied and accumulated with the image value $25$, then a weight-sharing \gls{mac} would compute $19 \times 25=475$ and add this value to the accumulator. Instead we keep $b$ separate accumulators, one for each weight value. If we encounter the shared weight with index $2$, value $19$ and image value $25$, then rather than performing any multiplication, we instead add $25$ to accumulator number $2$ in the local $b$-entry register file. Storing this result in a register file that is local the \gls{mac} unit reduces unnecessary data movement.

\begin{figure}[t]
	\centering
	\includegraphics[width=1\linewidth,keepaspectratio]{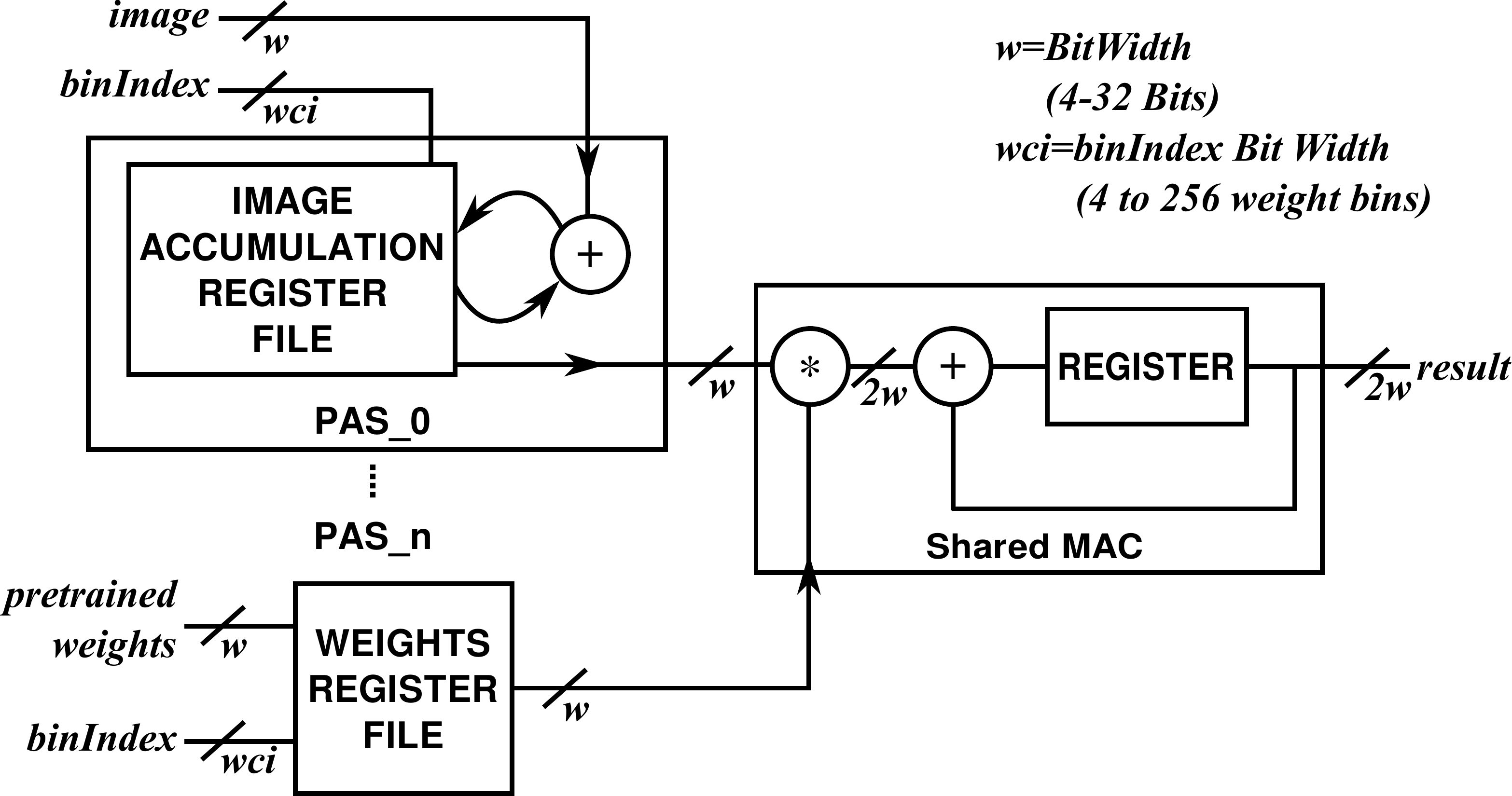}
	\setlength{\abovecaptionskip}{-0.25cm}
	\caption{Multiple-\gls{pas}-Shared-\gls{mac} showing \gls{pas} units followed by a shared \gls{mac} to multiply and accumulate \textbf{\textit{weights}} with binned \textbf{\textit{image}} values from bins for a bit width $w$ and a $wci$  \textbf{\textit{binIndex}} indexing into the $b$ weight-bin register. The accumulation phase accumulates \textbf{\textit{image}} values against a \textbf{\textit{binIndex}} representing how often the corresponding weight appears. The post-pass multiply-accumulate phase would multiply the weights with binned image values indexed by \textbf{\textit{binIndex}}.}
	\label{fig:pasmBlockDiagram}
\end{figure}

Our Parallel Accumulate and Store (\gls{pas}) unit represents the sum as an unevaluated sum of a set of coefficients of each of the $b$ weights (see Fig. \ref{fig:pasmBlockDiagram}). The system computes the dot product by computing the total of how many of each of the weights appear in the sum. This turns the multiply-accumulate step into an array-index-and-add operation.

To compute the final dot product value the final \gls{sop} of weights and their respective frequencies is calculated using a standard \gls{mac} unit. We refer to this combination of a \gls{pas} and \gls{mac} as a \gls{pasm}. The post-pass \gls{mac} requires just $b$ multiplications whereas the standard dot product needs as many multiplications as there are inputs. With far fewer multiplications needed, one post-pass \gls{mac} unit can be shared between several \gls{pas} units.

The \gls{pasm} proposed in Fig. \ref{fig:pasmBlockDiagram} stores the $w$ bit \textbf{\textit{image}} values in corresponding $b$ weight-bin registers indexed by the \textbf{\textit{binIndex}} signal. The \gls{pasm} has two phases: (1) to accumulate the image values into the weight bins (known as the \gls{pas}) and (2) to multiply the binned values with the weights (completing the \gls{pasm}).

\begin{figure}[t]
	\centering
	\includegraphics[width=0.60\linewidth]{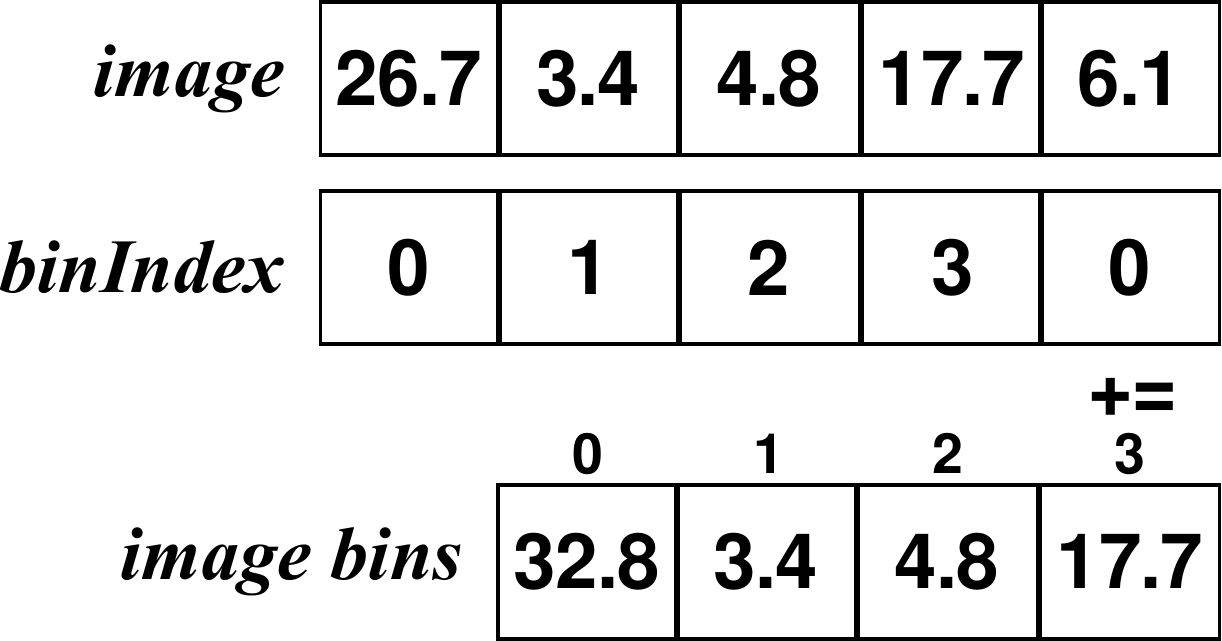}
	\setlength{\abovecaptionskip}{-0.01cm}
	\caption{Phase 1: As each image value is streamed in, its associated bin index is also streamed so that the image values can be accumulated into the correct bins.}
	\label{fig:pasPhase}
\end{figure}

Fig. \ref{fig:pasPhase} shows how \textbf{\textit{image}} value $26.7$ is accumulated into bin $0$ indexed by \textbf{\textit{binIndex}}. Next $3.4$ is accumulated into bin $1$. This continues until finally accumulating $6.1$ into bin $0$ to give $26.7+6.1=32.8$.

\begin{figure}[t]
	\centering
	\includegraphics[width=0.55\linewidth]{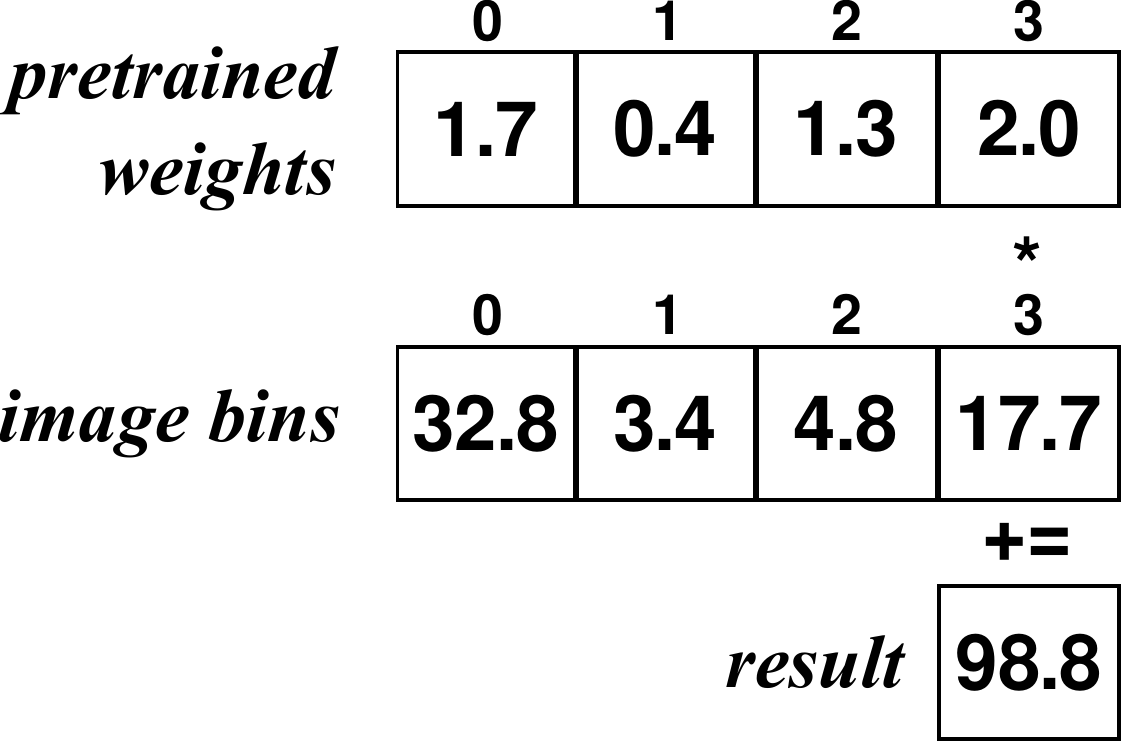}
	\setlength{\abovecaptionskip}{-0.01cm}
	\caption{Phase 2: Each accumulated value in each bin is multiplied with its corresponding pretrained weight value to produce the final equivalent result.}
	\label{fig:mulPhase}
\end{figure}

Fig. \ref{fig:mulPhase} demonstrates the multiply phase, multiplying and accumulating bin $0$ \textbf{\textit{pretrained weight}} with bin $0$ accumulated \textbf{\textit{image}} value, giving $32.8\times1.7=55.76$. The contents of \textbf{\textit{pretrained weight}} bin $1$ in multiplied and accumulated with \textbf{\textit{image}} value in bin $1$ and so on until all the corresponding bins are multiplied and accumulated into the \textbf{\textit{result}} register, giving $98.8$.

The latter \gls{mac} stage can be implemented on a \gls{mac} unit that is shared between several \gls{pas} units. \gls{mac} units can be replaced by \gls{pas} units and a number of \gls{pas} units share a single \gls{mac}. Due to its parallel nature, the \gls{pasm} can have higher throughput for known sets of weights when compared to the standard \gls{mac}.

Of course, from Table \ref{table:complexity}, the \gls{pasm} is only efficient in a weight shared \gls{cnn}. If this weight-shared accumulate-shared-multiply system were used in a standard \gls{cnn} then the number of the weight-bin registers would be prohibitive.

Table \ref{table:channelsKernelsMatrix} shows each sequence of \gls{mac} operations will consist of $k\times k\times c$ inputs (see lines 10 - 13 of Fig. \ref{fig:covCode}). For example, if $b=16$ bins is used for $k=5\times5$ kernels and $c=32$ channels, then 800 accumulations and 16 multipliers (one for each bin) would be required to do the accumulate and multiply. If $b=16$ bins is used, the overhead of the final accumulation is minimal for large numbers of kernels and channels. Careful consideration of the size of bins used with respect to the number of channels and kernels is important due to the $sum$ being multiply-accumulated $i$ many times before the $output$ is updated as can be seen on lines 10 - 13 of Fig. \ref{fig:covCode}.

\begin{table}[t]
	\centering
	\caption{Numbers of Channels Multiplied and Accumulated with the Kernels}
		\begin{tabular}{cllll}
		\multicolumn{1}{l}{}                  &                                   & \multicolumn{3}{c}{\textbf{input\_channels ($c$)}}                                                                \\ \cline{3-5} 
		\multirow{5}{*}{\textbf{kernels ($K$)}} & \multicolumn{1}{l|}{\textbf{}}    & \multicolumn{1}{l|}{\textbf{32}} & \multicolumn{1}{l|}{\textbf{128}} & \multicolumn{1}{l|}{\textbf{512}} \\ \cline{2-5} 
		& \multicolumn{1}{l|}{\textbf{1x1}} & \multicolumn{1}{l|}{32}          & \multicolumn{1}{l|}{128}          & \multicolumn{1}{l|}{512}          \\ \cline{2-5} 
		& \multicolumn{1}{l|}{\textbf{3x3}} & \multicolumn{1}{l|}{288}         & \multicolumn{1}{l|}{1152}         & \multicolumn{1}{l|}{4608}         \\ \cline{2-5} 
		& \multicolumn{1}{l|}{\textbf{5x5}} & \multicolumn{1}{l|}{800}         & \multicolumn{1}{l|}{3200}         & \multicolumn{1}{l|}{12800}        \\ \cline{2-5} 
		& \multicolumn{1}{l|}{\textbf{7x7}} & \multicolumn{1}{l|}{1568}        & \multicolumn{1}{l|}{6272}         & \multicolumn{1}{l|}{25088}        \\ \cline{2-5} 
	\end{tabular}
	\label{table:channelsKernelsMatrix}
\end{table}

\section{Evaluation}
\label{sec:evaluation}
We designed a multi-channel, multi-kernel convolution accelerator unit to perform a simplified version of the accumulations in Fig. \ref{fig:covCode}. Our accelerator accepts 4 image inputs and 4 shared-weight inputs each cycle, and uses them to compute 16 separate \gls{mac} operations each cycle. The standard version performs these operations on 16 weight-shared \gls{mac} units (16-\gls{mac}). Our proposed \gls{pasm} unit (16-\gls{pas}-4-\gls{mac}) has 16 \gls{pas} units and uses 4 \gls{mac} units for post-pass multiplication. Both are coded in Verilog 2001 and synthesized to a flat netlist at 100MHz with a short 0.1ns clock transition time targeted at a 45nm process \gls{asic}. We measure and compare the timing, power and gate count in both designs for the same corresponding bit widths and same numbers of weight bins.

The standard 16-\gls{mac} and the proposed 16-\gls{pas}-4-\gls{mac} each have $w$ bit \textbf{\textit{image}} and \textbf{\textit{weight}} inputs and the 16-\gls{pas}-4-\gls{mac} has a $wci$ bit \textbf{\textit{binIndex}} input to index into the $b=2^{wci}$ weight bins. The designs are coded using integer/fixed point precision numbers. Both versions are synthesized to produced a gate level netlist and timing constraints designed using \acrfull{sdc} \cite{FpgaXdcTiming:Gangadharan} so that both designs meet timing at 100MHz.

Cadence Genus (version 15.20 - 15.20-p004\_1) is used for synthesizing the \gls{rtl} into the OSU FreePDK 45nm process \gls{asic} and applying the constraints in order to meet timing. Genus supplies commands for reporting approximate timing, gate count and power consumption of the designs at the post synthesis stage. The ``report timing'', ``report gates'' and ``report power'' of Cadence Genus are used to obtain the results for both \gls{mac} and \gls{pasm} designs. Graphs of the gate count and power consumption results are produced for the two different designs at different bit widths and different numbers of weight bins, showing that the \gls{pasm} is consistently smaller and more efficient than the weight-sharing \gls{mac}.

\begin{figure}
	\centering
	\includegraphics[width=1\linewidth,keepaspectratio]{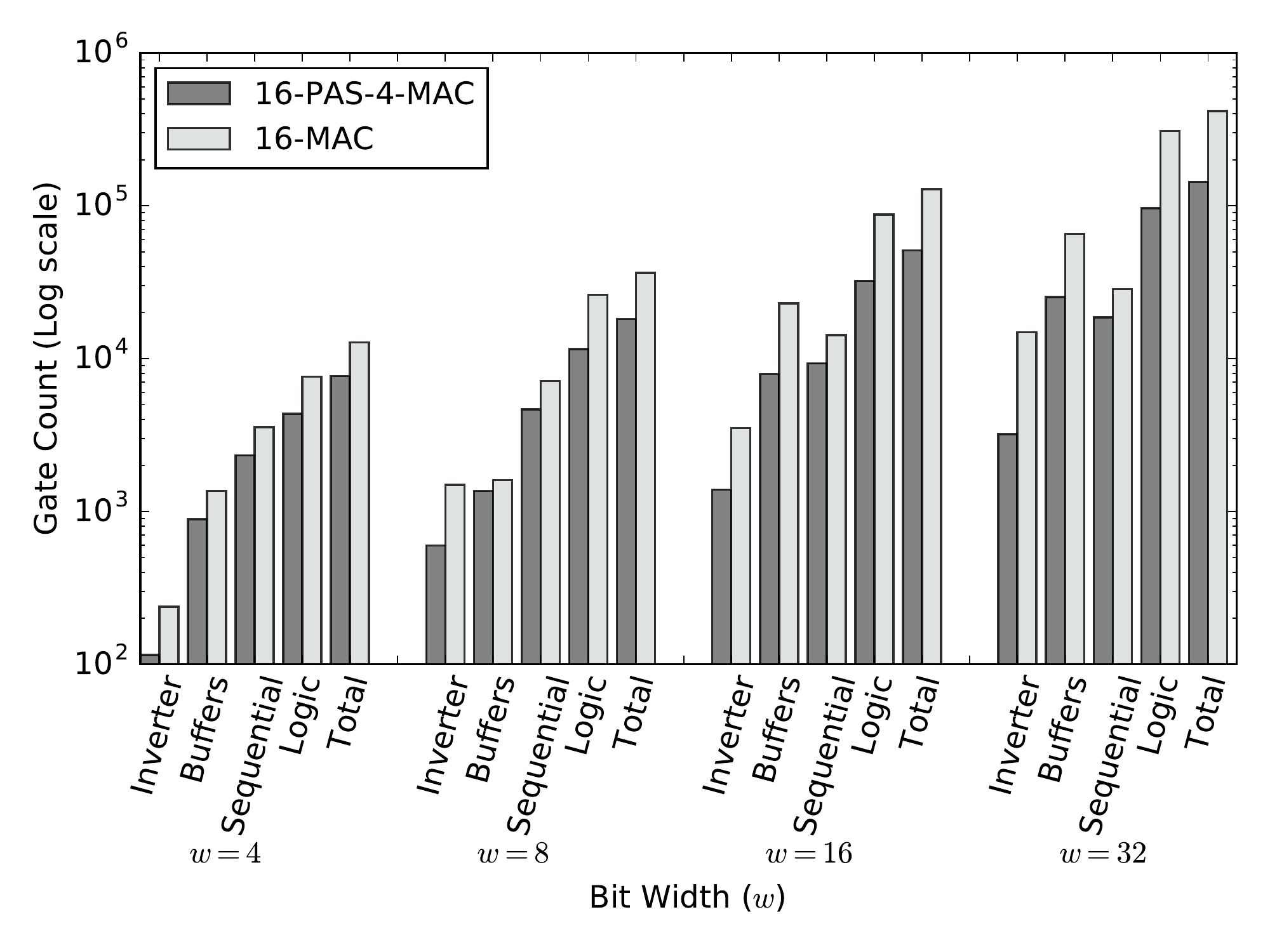}
	\setlength{\abovecaptionskip}{-0.65cm}
	\caption{Logic gate count comparisons (in NAND2X1 gates) for $w=4, 8, 16, 32$ bits wide 16-\gls{mac} and 16-\gls{pas}-4-\gls{mac} for $b=16$ weight bins - \textbf{lower is better} }
	\label{fig:asicUtilizationComparisonWidth}
\end{figure}

Fig. \ref{fig:asicUtilizationComparisonWidth} shows comparisons of the logic resource requirements of a $b=16$ shared-weight-bin 16-\gls{pas}-4-\gls{mac} and 16-\gls{mac} for varying $w$ bit widths. Gate counts are normalised to a NAND2X1 gate. The \gls{pasm} uses significantly fewer logic gates. For example, for $w=32$ bits wide the 16-\gls{pas}-4-\gls{mac} is 35\% smaller in sequential logic, 78\% smaller in inverters, 61\% smaller in buffers and 68\% smaller in logic, an overall 66\% saving in total logic gates. The \gls{pasm} requires more accumulators for the $b$-entry register file, but otherwise overall resource requirements are significantly lower than that of the \gls{mac}.

\begin{figure}
	\centering
	\includegraphics[width=1\linewidth,keepaspectratio]{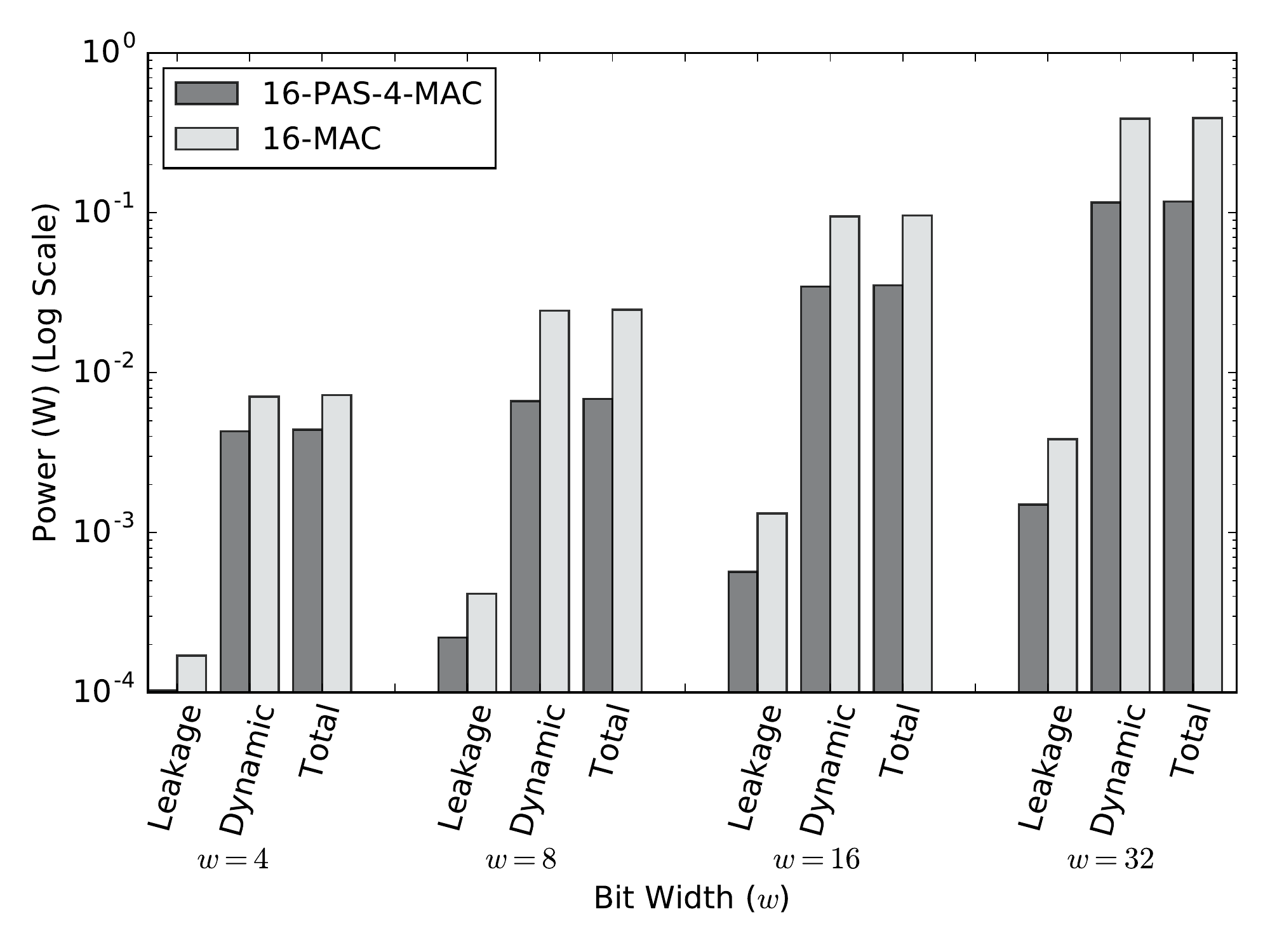}
	\setlength{\abovecaptionskip}{-0.65cm}
	\caption{Power consumption (in W) comparisons for $w=4, 8, 16, 32$ bits wide 16-\gls{mac} and 16-\gls{pas}-4-\gls{mac} for $b=16$ weight bins - \textbf{lower is better} } 
	\label{fig:asicTotalDynamicPowerComparisonWidth}
\end{figure}

Fig. \ref{fig:asicTotalDynamicPowerComparisonWidth} shows comparisons of power consumption of the designs. 16-\gls{pas}-4-\gls{mac}'s power is lower than the weight-shared 16-\glspl{mac} and the gap grows with increasing $w$ bit width. For example, for the $w=32$ bit versions of each design, the 16-\gls{pas}-4-\gls{mac} consumes 60\% less leakage power, 70\% less dynamic power and 70\% less total power than that of the 16-\glspl{mac}.

\begin{figure}
	\centering
	\includegraphics[width=1\linewidth,keepaspectratio]{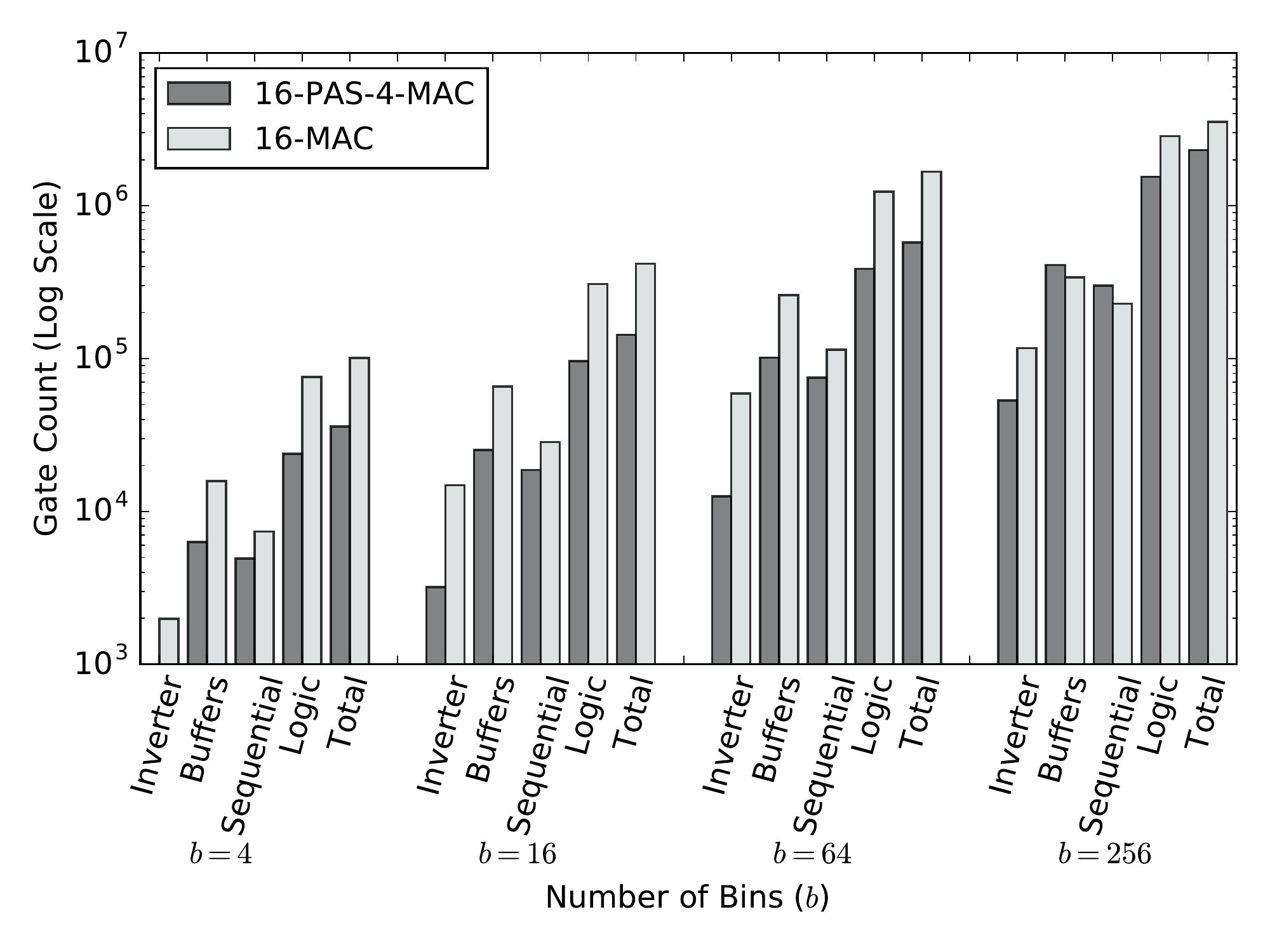}
	\setlength{\abovecaptionskip}{-0.65cm}
	\caption{Logic gate counts comparisons  (in NAND2X1 gates) for $b=4, 16, 64, 256$ weight bins for a 16-\gls{mac} and 16-\gls{pas}-4-\gls{mac} for $w=32$ bit width - \textbf{lower is better} }
	\label{fig:asicUtilizationComparisonDepth}
\end{figure}

Fig.  \ref{fig:asicUtilizationComparisonDepth} shows the effect of varying the number of bins from $b=4$ to $b=256$, with gate counts normalised to a NAND2X1. For bit width $w=32$ and $b=16$ bins the 16-\gls{pas}-4-\gls{mac} utilization has 35\% fewer sequential gates, 78\% fewer inverters, 62\% fewer buffers and 69\% fewer logic and 66\% less total logic gates compared to the 16-\gls{mac} design. However, at $b=256$, \gls{pasm} registers and buffers are less efficient than the \gls{mac}.

\begin{figure}[t]
	\centering
	\includegraphics[width=1\linewidth,keepaspectratio]{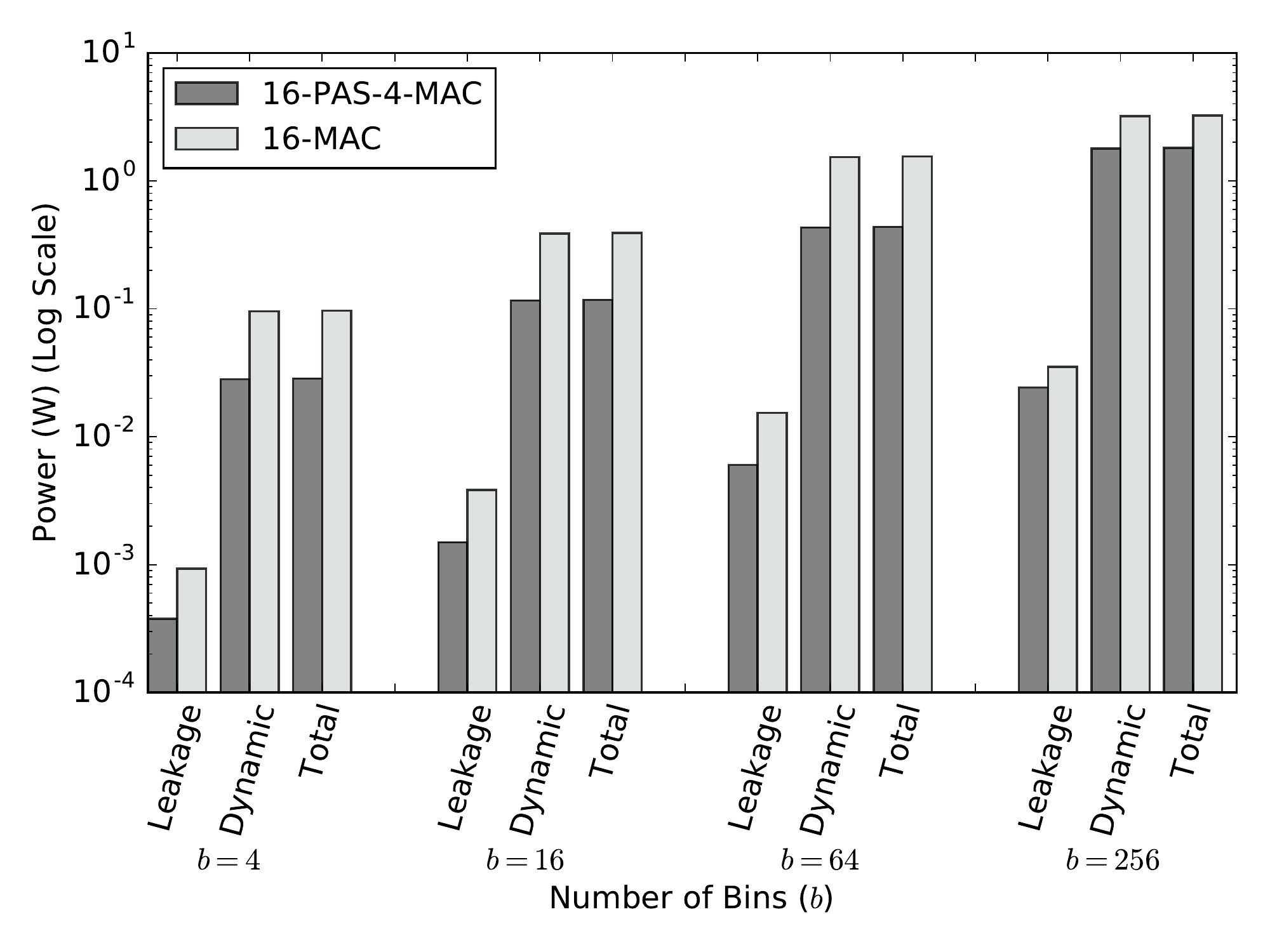}
	\setlength{\abovecaptionskip}{-0.65cm}
	\caption{Power consumption (in W) comparisons for $b=4, 16, 64, 256$ weight bins deep 16-\gls{mac} and 16-\gls{pas}-4-\gls{mac} for $w=32$ bit width - \textbf{lower is better} } 
	\label{fig:asicTotalDynamicPowerComparisonDepth}
\end{figure}

The 16-\gls{pas}-4-\gls{mac} also consumes 61\% less leakage power, 70\% less dynamic power and 70\% less total power (Fig. \ref{fig:asicTotalDynamicPowerComparisonDepth}).

We also experimented with implementing the designs on a Xilinx Kintex Ultrascale \gls{fpga} however, for larger values of $b$ bins, such as $b=64$ and $b=256$, our approach quickly becomes inefficient. 

The weight-shared \gls{pasm} introduces a delay in the processing the output of the \gls{pas} units. The \gls{pas} unit has a throughput of one pair of inputs per cycle, and so computes the initial accumulated values in about $n$ cycles. The post pass \gls{mac} unit also has a throughput of one pair of inputs per cycle, so requires a cycle for each of the $b$ accumulator bins, for a total of $n+b$ cycles. In contrast a simple \gls{mac} unit requires just $n$ cycles.

Whilst more parallelism of standard \glspl{mac} could be applied to accelerate a \gls{cnn}, these parallel \glspl{mac} would consume more power and area that that of our parallel \glspl{pas} and shared \glspl{mac}.

Future improvements should include placing and routing the \gls{mac} and \gls{pasm} designs to a fully routed test chip. Back annotated gate level simulations should be executed on the final netlist to obtain design toggle rates and applied to the power reporting tool. The placement and routing impacts and simulation toggle rates shall increase the power analysis accuracy comparisons of the designs.

\section{Related Work}
\label{sec:relatedWork}
Other research groups have proposed numerous hardware accelerators for \glspl{cnn} in both \gls{fpga} \cite{limitedNumericalPrecision:Gupta,OptimizingFpgaAccelForCNN:Zhang} and \gls{asic} \cite{DianNao:Chen}. Gupta \MakeLowercase{\textit{et al.}} \cite{limitedNumericalPrecision:Gupta} show increased efficiency in a hardware accelerator of a 16-bit fixed-point representation using stochastic rounding without loss of accuracy. Zhang \MakeLowercase{\textit{et al.}} \cite{OptimizingFpgaAccelForCNN:Zhang} deduce the best \gls{cnn} accelerator taking \gls{fpga} requirements into consideration and then implement the best on an \gls{fpga} to demonstrate high performance and throughput. Chen \MakeLowercase{\textit{et al.}} \cite{DianNao:Chen} design an \gls{asic} accelerator for large scale \glspl{cnn} focussing on the impact of memory on the accelerator performance.

Chen \MakeLowercase{\textit{et al.}} \cite{Eyeriss:Chen} address the problem of data movement which consumes large amounts of time and energy. They focus on data flow in the \gls{cnn} to minimize data movement by reusing weights within the hardware accelerator to improve locality. This was implemented in \gls{asic} and power and implementation results compared showing the effectiveness of weight reuse in saving power and increasing locality.

Han \MakeLowercase{\textit{et al.}} \cite{eie:Han} have proposed an Efficient Inference Engine which builds on their `Deep compression' \cite{DeepCompression:Han} work to perform inferences on the deeply compressed network to accelerate the weight-shared matrix-vector multiplication. This accelerates the classification task whilst saving energy when compared to \gls{cpu} or \gls{gpu} implementations.

Both weight sharing \cite{DeepCompression:Han,eie:Han} and weight reuse \cite{Eyeriss:Chen} reduce redundant data movement through different but complementary approaches. This paper builds on these ideas to reduce the circuit complexity of the underlying \gls{mac} units.

\section{Conclusion}
\label{sec:conclusion}
In this paper we propose a novel low-complexity \gls{mac} unit to exploit weight-sharing in \gls{cnn} accelerators. The gate count and area for the 16-\gls{pas}-4-\gls{mac} is significantly lower when compared with the 16-\gls{mac}. We also found better power efficiency with the 16-\gls{pas}-4-\gls{mac}. \gls{cnn} accelerators can contain hundreds or thousands of \gls{mac} units. We believe that our reduced-complexity design offers the possibility of adding more parallel \gls{mac} units to \gls{asic} hardware \gls{cnn} accelerators within a given clock speed, logic area and power budget.

\end{document}